\documentclass[5p]{elsarticle}
\usepackage{booktabs,shortvrb,url}
\usepackage{amsmath, amsfonts, amssymb}
\usepackage{algpseudocode, algorithm}
\makeatletter\def\ps@pprintTitle{\let\@oddhead\@empty\let\@evenhead\@empty\let\@oddfoot\@empty\let\@evenfoot\@oddfoot}
\makeatother
\begin{document}
\begin{frontmatter}
\title{Soft Computing approaches on the Bandwidth Problem}
\author[1]{Gabriela Czibula}\ead{gabis@cs.ubbcluj.ro}
\author[2]{Gloria Cerasela Cri\c san}\ead{ceraselacrisan@ub.ro}
\author[3]{Camelia-M. Pintea}\ead{cmpintea@yahoo.com}
\author[1]{Istvan-Gergely Czibula}\ead{istvanc@cs.ubbcluj.ro}
\address[1]{Department of Computer Science, Babe\c s-Bolyai University, Cluj-Napoca, Romania}
\address[2]{Faculty of Science, "Vasile Alecsandri" University of Bac\u au, Romania}
\address[3]{Technical University Cluj Napoca, North University Center Baia Mare, Romania}
\begin{abstract}
The Matrix Bandwidth Minimization Problem (\emph{MBMP}) seeks for a simultaneous reordering of the rows and the columns of a square matrix such that the nonzero entries are collected within a band of small width close to the main diagonal. The \emph{MBMP} is a NP-complete problem, with applications in many scientific domains, linear systems, artificial intelligence, and real-life situations in  industry, logistics, information recovery.
The complex problems are hard to solve, that is why any attempt to improve their solutions is beneficent. \emph{Genetic algorithms} and \emph{ant-based systems} are Soft Computing methods used in this paper in order to solve some \emph{MBMP} instances. Our approach is based on a learning agent-based model involving a local search procedure. The algorithm is compared with the classical {\em Cuthill-McKee} algorithm, and with a hybrid genetic algorithm, using several instances from {\em Matrix Market} collection. Computational experiments confirm a good performance of the proposed algorithms for the considered set of \emph{MBMP} instances. On {\em Soft Computing} basis, we also propose a new theoretical \emph{Reinforcement Learning} model for solving the \emph{MBMP} problem.
\end{abstract}

\end{frontmatter}

\section{Introduction}\label{sec:intro}
\emph{Combinatorial optimization} is the seeking for one or more optimal solutions in a well defined discrete problem space. In real life approaches, this means that people are interested in finding efficient allocations of limited resources for achieving desired goals, when all the variables have integer values. As workers, planes or boats are indivisible (like many other resources), the Combinatorial Optimization Problems (COPs) receive today an intense attention from the scientific community.

The current real-life COPs are difficult in many ways: the solution space is huge, the parameters are linked, the decomposability is not obvious, the restrictions are hard to test, the local optimal solutions are many and hard to locate, and the uncertainty and the dynamism of the environment must be taken into account. All these characteristics, and other more, constantly make the algorithm design and implementations challenging tasks. The quest for more and more efficient solving methods is permanently driven by the growing complexity of our world. 

The {\em Matrix Bandwidth Minimization Problem (MBMP)} is a fundamental mathematical problem, searching for a simultaneous permutation of the rows and the columns of a square matrix that keeps its nonzero entries as much as possible close to the main diagonal. This problem is NP-complete in general \cite{Papadimitriou1976}, and it remains so even in restricted solutions spaces \cite{Garey1978}, that is why any attempt to improve its solutions is beneficent.

The main contribution of this paper is to emphasize the effectiveness of using \emph{soft computing} methods in order to solve the \emph{Matrix Bandwidth Minimization Problem}. \emph{Genetic algorithms} and \emph{ant-based systems} are natural computing methods used in this paper in order to solve the \emph{MBMP} instances.  Computational experiments confirm that these methods provide robust and low-cost solutions for the \emph{MBMP}. We also introduce a new theoretical \emph{reinforcement learning} model for solving the \emph{MBMP}. So far, such a learning model has not been reported in the \emph{MBMP} literature.

The rest of the paper is organized as follows. Section \ref{mbmp} briefly presents the matrix bandwidth minimization problem, emphasizing its relevance and also presenting existing approaches for solving it. The fundamentals of the soft computing approaches considered in this paper, i.e genetic algorithms, ant colony systems and reinforcement learning, are given in Section \ref{bck}. In Section \ref{natcomp} we propose two \emph{natural computing} methods for solving the \emph{MBMP} instances, namely \emph{genetic algorithms} and \emph{ant colony systems}. A theoretical reinforcement learning model for solving \emph{MBMP} is introduced in Section \ref{prop}. Section \ref{exp} provides an experimental evaluation of the proposed methods and Section \ref{conc} contains some conclusions of the paper and future development of our work.

\section{The Matrix Bandwidth Minimization Problem}\label{mbmp}
This section introduces the concept and the literature review related to the  \emph{Matrix Bandwidth Minimization Problem}.
\subsection{Matrix Bandwidth Minimization Problem description}\label{mbmp1}\medskip
\noindent Given a square positive symmetric matrix $\mathcal{A}=(a_{ij})_{1\le i,j\le n}$ the bandwidth $\beta $ is the value $\beta(\mathcal{A})=\mathop{max}_{a_{ij}\neq 0}|i-j|$. The Matrix Bandwidth Minimization Problem searches for a row (and column) permutation $\pi$ that minimizes the bandwidth for the new matrix.

An equivalent form of \emph{MBMP} uses the graph-theory approach, based on the \emph{layout} notion. Given an undirected, connected graph \emph{G=(V, E)}, a layout $\sigma$  is a bijection between \emph{V} and $\{1, 2,\dots |V|\}$. The bandwidth of G is $\beta(G)=min_\sigma$ $(max_{(u,v)\in V}(|\sigma(u)-\sigma(v)|))$. Intuitively, computing the bandwidth for a graph is to find a linear ordering of its vertices that minimizes the maximum distance between two adjacent vertices.

Starting from the given matrix $\mathcal{A}$, an equivalent graph $G_\mathcal{A}=(V,E)$ can be defined and the \emph{MBMP} can be viewed as the problem of minimizing the bandwidth of $G_\mathcal{A}$. In this graph, the set of vertices is $V=\{1,2,\dots,n\}$ and two vertices $i$ and $j$ are connected through an edge {\it iff} $a_{ij} \neq 0$, i.e $E=\{(i,j)\;${\it iff}$\;a_{ij}\neq 0 \}$.

The current exact approaches devise algorithms that solve the general \emph{MBMP} in $O(4.83^n)$  running time \cite{Cygan2009}. Classic results for approximation approaches establish an approximation factor of $O(log^{3.5}n)$ for general \emph{MBMP} \cite{Feige1998} and $O(log^{2.5}n)$ for trees \cite{Gupta2000}.

The {\it MBMP} arose in the solving systems of linear equation; the ordering of the system matrix has great impact on the resources needed when actually solving the system, and may lead to a substantial efficiency increase. Minimizing the bandwidth of a matrix helps in improving the efficiency of certain linear algorithms, like Gaussian elimination.

The \emph{MBMP} current applications in computer science include VLSI design, network survivability, data storage. Other applications are in electromagnetic industry \cite{Esposito1998}, large-scale power transmission systems, chemical kinetics and numerical geophysics \cite{Marti2001}, information retrieval in hypertext \cite{Berry1996}.

Some generalizations of \emph{MBMP} are currently investigated by the world researchers. For example, the two-dimensional bandwidth problem is to embed a graph into a planar grid such that the maximum distance between adjacent vertices is as small as possible \cite{Lin2010}.

\subsection{Literature review.}\medskip 
The importance of the bandwidth minimization problem is also reflected by the large number of publications describing algorithms for solving it. Cuthill and McKee propose in 1969 in \cite{Cuthill1969} the first stable heuristic method for \emph{MBMP}: the \emph{CM} algorithm with {\it Breadth-First Search}.  

Marti et al. have used in \cite{Marti2001} Tabu Search for solving the \emph{MBMP} problem. They
used a candidate list strategy to accelerate the selection of moves in the neighborhood of the current solution. 

A $GRASP$ with \emph{Path Relinking} method given by Pinana et al. in \cite{Pinana2004} has been shown to achieve better results than the Tabu Search procedure but with longer running times. Lim et al. propose in \cite{Lim2003} a Genetic Algorithm integrated with Hill Climbing to solve the bandwidth minimization problem. 

A simulated annealing algorithm is shown in \cite{Rodriguez-Tello2008} for the matrix bandwidth minimization problem. The algorithm proposed by Tello et al. is based on three distinguished features including an original internal representation of solutions, a highly discriminating evaluation function and an effective neighborhood. More recently, the {\it Ant Colony Optimization} metaheuristic has been used in \cite{Lim2006}, \cite{Pintea2010} in order to solve the the \emph{MBMP}.

\section{Background}\label{bck}

In this section we will briefly review the fundamentals of the soft computing approaches used in this paper for solving the \emph{MBMP}, i.e \emph{genetic algorithms}, \emph{ant colony optimization} and \emph{reinforcement learning}.

{\em Soft Computing} is the collection of computing branches that cope with the imprecision, uncertainty, partial truth, and approximation, manifested in nature and naturally (and gracefully) operated by biologic entities (cells, organisms, or collections of individuals). The goal of soft computing approaches is to achieve tractability, robustness and low-cost solutions, facing the real-life, complex, highly-dimensioned problems.

\emph{Genetic algorithms} (GAs), invented by John Holland in the 1960s, are the most widely used approaches to computational evolution.  Genetic algorithms provide an approach to machine learning \cite{Mitchell1998}, method motivated by analogy to biological evolution. Hypotheses are often described by bit strings whose interpretation depends on the application, though hypotheses may also be described by symbolic expressions or even computer programs \cite{Goldberg1989}. 

{\it Ant Colony Optimization (ACO)} studies artificial systems inspired by the behavior of real ant colonies and which are used to solve COPs \cite{Dorigo2004}. The {\it ACO} methods use a set of cooperative artificial ants, each constructing a solution, based on the expected quality of the available moves and on the good solutions found by the community. {\it ACO} demonstrated a high flexibility and strength by solving with very good results either academic instances of many COPs or real-life problems. 
To improve the efficiency, the ant-based algorithms are designed using problem-specific information and involve local search methods.
 
The goal of building systems that can adapt to their environments and learn from their experiences has attracted researchers from many fields including computer science, mathematics, cognitive sciences \cite{Sutton1998}. 

\emph{Reinforcement learning} (RL) is learning what to do - how to map situations to actions - so as to maximize a numerical \emph{reward} signal. The learner is not told which actions to take, as in most forms of machine learning, but instead must discover which actions yield the highest reward by trying them. In RL, the computer is simply given a goal to achieve. The computer then learns how to achieve that goal by trial-and-error interactions with its environment.

\section{Natural computing models for the MBMP}\label{natcomp}

In this section we propose two \emph{natural computing} methods for solving the \emph{MBMP} instances: \emph{genetic algorithms} and \emph{ant colony systems}. The computational experiments from Section \ref{exp} confirm that these methods provide robust and low-cost solutions for the \emph{MBMP}.

\subsection{Genetic Algorithm.}\label{ga}

In the following, a hybrid genetic algorithm (\emph{HGA}) is proposed for solving the \emph{MBMP}. The algorithm proposed in this section is a slight modification of the Genetic Algorithm integrated with Hill Climbing proposed by Lim et al. in \cite{Lim2003}.

Let us consider that $\mathcal{A}=(a_{ij})_{1\le i,j\le n}$ is the square symmetric matrix whose bandwidth $\beta$ has to be minimized.

In the \emph{HGA} we use, a chromosome is a $n$ dimensional sequence $\pi_1,\pi_2,\dots\pi_n$ representing a permutation $\pi$ of \\$\{1,2,\dots,n\}$.
Thus, a matrix $\mathcal{A}_\pi$ can be associated to a chromosome $\pi$, i.e the matrix obtained starting from the matrix $\mathcal{A}$ by permuting it rows (and columns) in the order given by permutation $\pi$. The fitness function associated to a chromosome $\pi$ is defined as the bandwidth of the corresponding matrix, i.e $fitness(\pi)=\beta(\mathcal{A}_\pi)$. The problem consists of minimizing the fitness function, i.e finding the individual with the minimum associated fitness value.

We have used the traditional structure for a genetic algorithm, adding a Hill Climbing step in order to quickly tune solutions to reach local optimum \cite{Lim2003}. \emph{HGA} algorithm operates as follows:
\begin{itemize}
\item [i.] At the beginning, an initial
group of $n$ chromosomes is constructed, as it will be further detailed.
\item [ii.] Then, middle-point crossover and a $k$-swap mutation are performed on this group of chromosomes to generate new chromosomes \cite{Lim2003}. Hill Climbing is now applied to each newly-generated chromosome, as proposed in \cite{Lim2003}.
As the number of individuals within a population remains $n$, fittest chromosomes will remain in the next generation. After the new generation is formed, a swap mutation is applied on all chromosomes within the new generation, excepting the best one. Then, Hill Climbing is applied again to each newly-generated chromosome.
\item [iii.] Step [ii.] is repeated for a given number of generations; the algorithm stops and the best result is reported as solution.
\end{itemize}
The initial population for the \emph{HGA} is constructed as follows. Starting from matrix $\mathcal{A}$, we construct the corresponding graph $G_\mathcal{A}$. Then, the initial chromosomes are built by performing BFS on the graph, starting from each node. This way, $n$ initial individuals are constructed. Applying Hill Climbing \cite{Lim2003} on the obtained individuals, the initial population for the \emph{HGA} is obtained. The construction of the initial population for the \emph{HGA} is slightly different from the method from \cite{Lim2003}.

As further work we will investigate the appropriateness of replacing the Hill Climbing step from the \emph{HGA} with other local search mechanisms, such as {\it PSwap} or {\it MPSwap} procedures that will be described in Subsection \ref{ant}.

\subsection{Ant-based system.}\label{ant}

A hybridized ACO approach using a local search procedure is proposed in this section for solving the \emph{MBMP}. This local search method is designed to reduce the bandwidth of the current solution and is executed during the local search stage of the ACO framework.

In \cite{Pintea2010} {\it Ant Colony System (ACS)} \cite{Dorigo2004} is hybridized with a local search mechanism. The {\it ACS} model is based on the level structure used by the Cuthill-McKee algorithm \cite{Cuthill1969}. The local search procedure aims at improving {\it ACS} solutions, by reducing the maximal bandwidth. The integration of a local search phase within the proposed {\it ACS} approach to \emph{MBMP} facilitates the refinement of ants' solutions.

The main stages of the proposed hybrid {\it ACS} are as follows.

\begin{itemize}
\item [i.] First, the current matrix bandwidth is computed, the pheromone trails are initialized and the parameters values are established.
\item [ii.] The construction stage consists of executing the next steps within a given number of iterations. At first all the ants are placed in the node from the first level, then the local search mechanism is applied. 

Each ant builds a feasible solution by repeatedly making pseudo-random choices from the available neighbors. While constructing its solution, an ant also modifies the amount of pheromone on the visited edges by applying the local updating rule \cite{Dorigo2004}. 

After each partial solution is built, in order to improve each ant's solution, the local search mechanism is applied. Finally, once all ants have finished  their tour, the amount of pheromone on edges is modified again by applying the global updating rule \cite{Dorigo2004}. 
\item [iii.] The best current solution is listed.
\end{itemize}

As illustrated above, the local search procedure is used twice within the proposed hybrid model: at the beginning of each iteration and after each partial solution is built, in order to improve each ant's solution.

In \cite{Pintea2010} two local search mechanisms are introduced: {\it PSwap} and {\it MPSwap}. The local search mechanisms are denoted by {\it hACS} and respectively {\it hMACS}. 

{\it PSwap} firstly founds the maximum and minimum degrees. Then, for all indexes $x$ with the maximum degree, it randomly selects an unvisited node $y$ with a minimum degree and then swaps the nodes $x$ and $y$.

In order to avoid stagnation was introduced {\it hMACS}. First are found the maximum and minimum degrees. For all indexes $x$ with the maximum degree, it randomly selects an unvisited node $y$ with a minimum degree such as the matrix bandwidth decreases and then swaps the nodes $x$ and $y$.

The experimental results reported in \cite{Pintea2010} show that {\it MPSwap} procedure performs better on small instances, while {\it PSwap} is better on larger ones.

\section{A theoretical reinforcement learning model for solving MBMP}\label{prop}

In this section we investigate a reinforcement learning approach for solving the \emph{MBMP} problem and introduce our RL model.

Let us assume, in the following, that $\mathcal{A}$ is the symmetric matrix of order $n$ whose bandwidth has to be minimized.

\subsection{Problem definition.}

\noindent We define the RL problem associated to \emph{MBMP} as in \cite{Czibula2010}:

\begin{itemize}
\item The environment $E$ consists of the set of states \\ $\{1,2,\dots,n\}$ extended with a state $s_0$ that is connected to all other states,  i.e. $E=\{1,2,\dots,n\} \bigcup \{s_0\}$.
\item The initial state $si$ of the agent in the environment is $s_0$.
\item A state $sf \in E$ reached by the agent at a given moment after it has visited states $si, s_1, s_2,\dots s_k$ is a \emph{terminal} (final) state if the number of states visited by the agent in the current sequence is $n+1$, i.e. $k=n$.
\item The transition function between the states is defined as $h:E \rightarrow P(E)$, where $h(i)=\{1,2,\dots n\}$. This means that, at a given moment, from the state {\it i} the agent can move to any state from $E$, excepting the initial state. We say that a state {\it j} that is accessible from state {\it i} ($j \in h(i)$) is the {\it neighbor} ({\it successor}) state of {\it i}.
\item The transitions between the states are equiprobable, the transition probability $P(i,j)$ between a state {\it i} and each neighbor state {\it j} of {\it i} is $P(i,j)=\frac{1}{n}$.
\end{itemize}

The RL problem consists in training the \emph{MBMP} agent to find a path $si, \pi_1, \pi_2,\dots \pi_n$ from the initial to a final state, i.e a permutation $\pi$ of $\{1,2,\cdots,n\}$ that minimizes the corresponding matrix bandwidth \cite{Czibula2010}. Let us consider that, at a given moment, the agent has visited states $si, \pi_1, \pi_2,\dots \pi_k$, where $k \le n$, $\pi_i\in E$, $\pi_i \neq si$; $\forall$ $1\le i\le k$ and $\pi_i\neq \pi_j$, $\forall$ $1\le i,j\le k, i\neq j$. 
Starting from the path $\pi_1, \pi_2,\dots \pi_k$, we construct a permutation of $\{1,2,\dots,n\}$, denoted by $\sigma^{\pi(1..k)}=(\sigma^{\pi(1..k)}_1, \sigma^{\pi(1..k)}_2,\dots,\sigma^{\pi(1..k)}_n)$. An element $\sigma^{\pi(1..k)}_j$ ($\forall 1 \le j \le n$) is computed as follows:

\begin{itemize}
\item[-] If $j \le k$, then $\sigma^{\pi(1..k)}_j=\pi_j$.
\item[-] If $k < j \le n$  and  $j \notin \{\pi_1,$  $\pi_2,$ $\dots$ $\pi_k\}$, then \\$\sigma^{\pi(1..k)}_j=j$.
\item[-] If $k < j \le n$  and  $j \in \{\pi_1,$  $\pi_2,$ $\dots$ $\pi_k\}$, then \\$\sigma^{\pi(1..k)}_j=s$, where $1 \le s \le n$, $ s \notin \{\pi_1, \pi_2,\dots\pi_k\}$ and $\exists m, \; 1 < m < k$ and $i_1, i_2,\dots,i_m \; (1 \le i_q \le n\;$ $ \forall$   $1 \le q \le m)$ such that $j=\pi_{i_1}$,  $i_1=\pi_{i_2},  \dots,$  $i_{m-1}=\pi_{i_m}$, $i_m=\pi_s$.
\end{itemize}

Based on the definition of $\sigma^{\pi(1..k)}$ given above, it can be proved that $\sigma^{\pi(1..k)}$ is a permutation of $\{1,2,\dots,n\}$. Now, a matrix $\mathcal{A}^{\sigma^{\pi(1..k)}}$ can be obtained from the initial matrix $\mathcal{A}$ by permuting its rows (and columns) in the order given by permutation $\sigma^{\pi(1..k)}$.

Consequently, a path $\pi_1, \pi_2,\dots\pi_k$ of the agent in the environment corresponds to the matrix $\mathcal{A}^{\sigma^{\pi(1..k)}}$ obtained as we have described above.

\subsection{Reinforcement function.}

As we aim at obtaining a permutation $\pi$ of $\{1,2,\dots,n\}$ that minimizes the matrix bandwidth, we define the reinforcement function as indicated in Equation (\ref{eq}). We mention that an alternative method to define the reinforcement function was considered in \cite{Czibula2010}. 

\begin{itemize}
\item the reward received in state $\pi_k$ after states $si,$ $\pi_1,\pi_2,\dots$  $\pi_{k-1}$ were visited, denoted by $r(\pi_k|si, \pi_1, \dots\ pi_{k-1})$ is computed as the bandwith of matrix $\mathcal{A}^{\sigma^{\pi(1..k-1)}}$ minus the bandwidth of matrix $\mathcal{A}^{\sigma^{\pi(1..k)}}$.
\end{itemize}

\begin{equation}\label{eq}
r(\pi_k|si,\pi_1,\dots\pi_{k-1})=\left\{\begin{array}{ll}
0&\mbox{if }k=1\\
\beta(\mathcal{A}^{\sigma^{\pi(1\dots k-1)}})-\beta(\mathcal{A}^{\sigma^{\pi(1\dots k)}})&\mbox{otherwise}\\
\end{array}, \right.
\end{equation}

Considering the reward defined in Equation (\ref{eq}), as the learning goal is to maximize the total amount
of rewards received on a path from the initial to a final state, it can be easily proved that the agent is trained to find a permutation $\pi$ of $=\{1,2,\dots,n\}$ that minimizes the  bandwidth of the corresponding matrix $\mathcal{A}^{\sigma^{\pi(1..k)}}$.

\subsection{The learning process.}\medskip

During the training step of the learning process, the agent will determine its \emph{optimal policy} in the environment, i.e the \emph{policy} that maximizes the sum of the received rewards. During the training process, the states' utilities estimations converge to their exact values, thus, at the end of the training process, the estimations will be in the vicinity of the exact values.

It is proved that the RL algorithm (such as SARSA \cite{Sutton1998}) converges with probability 1 to an utility function as long as all state-action pairs are visited an infinite number of times and the policy converges in the limit to the Greedy policy. 

Consequently, after the training step of the agent has been completed, the solution learned by the agent will be constructed by starting from the initial state and following the $Greedy$ policy until a solution is reached. From given state $i$, using the $Greedy$ policy, the agent moves to an unvisited neighbor $j$ of $i$ having the maximum utility value. The solution of the \emph{MBMP} reported by the RL agent is a permutation $\pi$ of $\{1,2,\dots,n\}$ such that $U(\pi_1) \ge U(\pi_2) \ge \dots \ge U(\pi_n)$, $U$ being the utility function learned by the agent during its training. Considering the general goal of a RL agent, it can be proved that the permutation $\pi$ of $\{1,2,\dots,n\}$ learned by the $MBMP$ agent converges to the permutation that corresponds to the matrix with the minimum bandwidth. \medskip

\section{Computational experiments}\label{exp}\medskip

In this section follows the comparative evaluation of the techniques proposed in Section \ref{natcomp} in order to solve the \emph{MBMP}. The results are compared with those reported by {\it CM} algorithm \cite{Cuthill1969}.

Nine benchmark instances from \textit{National Institute of Standards and Technology, Matrix Market, Harwell-Boeing sparse matrix collection} \cite{Harwell-Boeing2010} were used in the  computational experiments. In Table \ref{table:data} are illustrated, for each considered instance, the following characteristics: number of lines, number of columns and number of nonzero entries.

\begin{table}[ht]
\caption{The benchmark instances.}\medskip \centering
\begin{tabular}{c c c}\hline\hline
No. & Instance & Euclidean Characteristics \\\hline
1 & can\_24  & 24 24 92    \\
2 & can\_61  & 61 61 309   \\
3 & can\_62  & 62 62 140   \\
4 & can\_73  & 73 73 225   \\
5 & can\_96  & 96 96 432   \\
6 & can\_187 & 187 187 839 \\
7 & can\_229 & 229 229 1003\\
8 & can\_256 & 256 256 1586\\
9 & can\_268 & 268 268 1675\\\hline
\end{tabular}\label{table:data}
\end{table} \bigskip

The hybrid genetic algorithm {\it HGA} (Subsection \ref{ga}) and the hybrid ant systems {\it hACS} and {\it hMACS} (Subsection \ref{ant}) were implemented and  applied for the instances described in Table \ref{table:data}. Some details regarding the implementations of {\it HGA}, {\it hACS} and {\it hMACS} are following. 

The {\it Hybrid GA} is based on a {\it Delphi} implementation \cite{Zavoianu2009} and is tested with $10\%$ mutation rate, $k=[n/10]$ and $50$, respectively $100$ generations. $GA1$ is denoted the hybrid genetic algorithm with $50$ generations and $GA2$ the hybrid genetic algorithms with $100$ generations.

The hybrid ant algorithms \cite{Pintea2010} {\it hACS} and {\it hMACS} are implemented in Java. For each instance, both algorithms are executed $20$ times. 

The parameter values for both implementations are: $10$ ants, $10$ iterations, $q_0=0.95$, $\beta=2$, $\rho=0.001$, $\tau_0=0.1$. The algorithms were compiled on an AMD 2600 computer with 1024 MB memory and 1.9 GHz CPU clock.

In Table \ref{table:sol} are comparatively illustrated the best solution (the bandwidth of the matrix) obtained by {\it CM}, {\it hACS}, {\it hMACS},  {\it GA1} and {\it GA2} algorithms for the instances given in Table \ref{table:data}.

\begin{table}[h]
\caption{Comparative results.}\medskip\centering
\begin{tabular}{cccccc}\hline\hline
Instance no.& CM & hACS&hMACS&GA1&GA2\\\hline
1 & 8          & 14          & 11           & \textbf{6}  & \textbf{6}  \\
2 & 26         & 43          & 42           & \textbf{19} & \textbf{19} \\
3 & 9          & 20          & 12           & \textbf{8}  & \textbf{8}  \\
4 & 27         & 28          & \textbf{22}  & \textbf{22} & 23          \\
5 & 23         & \textbf{17} & \textbf{17}  & 25          & 25          \\
6 & \textbf{23}& 63          & 33           & 53          & 51          \\
7 & \textbf{49}& 120         & 120          & 63          & 63          \\
8 &116         & 148         & 189          & \textbf{91} & \textbf{91} \\
9 &134         & 165         & 210          & \textbf{90} & \textbf{90} \\\hline
\end{tabular}\label{table:sol}
\end{table}\medskip

\begin{figure}[htbp] \includegraphics[scale=0.3]{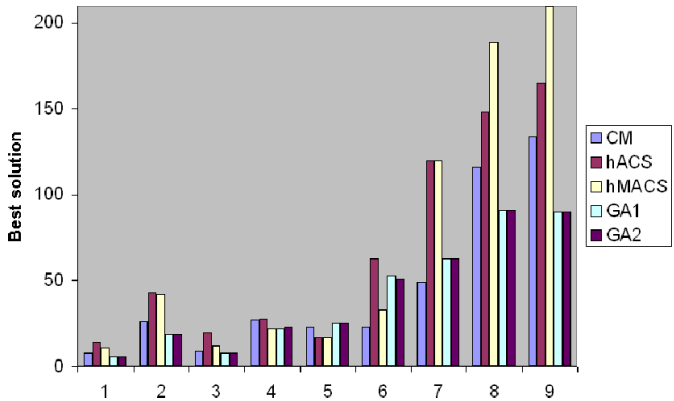}\caption{Comparative results.}\label{fig}\end{figure}

A graphical representation of the results is given in Figure \ref{fig}. Based on Figure~\ref{fig} some conclusions follows. 

Excepting two instances ($6$ and $7$) the hybrid natural-based algorithms provide better result than {\it CM} algorithm. {\it hMACS} algorithm performs better than {\it hACS} algorithm on small instances, while {\it hACS} algorithm is better than {\it hMACS} on larger ones. For six instances the hybrid genetic algorithm performed better than ant-based algorithms. The number of generations considered for {\it GA1} and {\it GA2} has no significant influence on the results. 

In order to assure a better convergence to the solution, the ant-based hybrid models should offer an "ideal" set of parameters and also a good strategy of placing the agents in the environment.  

The {\it Matrix Bandwidth Minimization Problem}'s results could be improved using reinforcement learning in new hybrid natural based-computing techniques.\medskip

\section{Conclusions and further work}\label{conc}

The \emph{Matrix Bandwidth Minimization Problem (MBMP)} is a classic mathematical problem, relevant to a wide range of complex real life applications. The problem is  NP-complete and a lot of research was conducted in order to find appropriate solutions.

Nowadays, bio-inspired heuristics are successfully used to solve difficult pro\-blems.  On this basis, the paper describes several soft computing approaches for solving the \emph{MBMP}. The proposed heuristics are hybrid algorithms: genetic algorithms and ant colony algorithms. Some standard \emph{MBMP} instances are tested using the hybrid bio-inspired algorithms and compared with existing literature. The results are encouraging. 

A new theoretical \emph{reinforcement learning} model for solving the considered problem is also introduced. Computational experiments confirmed a good performance of the proposed algorithms, emphasizing the effectiveness of \emph{soft computing} methods in order to solve the \emph{MBMP}.

Further work will be made in order to detail the proposed reinforcement learning model. More exactly, we proposed to develop a {\it RL} algorithm for training the \emph{MBMP} agent and to experimentally validate the {\it RL} model. We will also investigate new local search procedures in order to improve the performance of the ant system and of the genetic algorithm proposed for solving the \emph{Matrix Bandwidth Minimization Problem}.

\end{document}